# Exploiting Capacity of Sewer System Using Unsupervised Learning Algorithms Combined with Dimensionality Reduction

*Duo Zhang; Geir Lindholm; Nicolas Martinez; Harsha Ratnaweera*

**Abstract:** Exploiting capacity of sewer system using decentralized control is a cost-effective mean of minimizing the overflow. Given the size of the real sewer system, exploiting all the installed control structures in the sewer pipes can be challenging. This paper presents a 'divide and conquer' solution to implement decentralized control measures based on unsupervised learning algorithms. A sewer system is first divided into a number of sub-catchments. A series of natural and built factors that have the impact on sewer system performance is then collected. Clustering algorithms are then applied to grouping sub-catchments with similar hydraulic/hydrologic characteristics. Following which, principal component analysis is performed to interpret the main features of sub-catchment groups and identify priority control locations. Overflows under different control scenarios are compared based on the hydraulic model. Simulation results indicate that priority control applied to the most suitable cluster could bring the most profitable result.

**Keywords:** sewer control; machine learning; hydraulic model; clustering

**Authors:**

Duo Zhang:

    Ph.D. candidate, Faculty of Science and Technology, Norwegian University of Life Sciences, 1432, Ås, Norway.

Geir Lindholm:

    CEO of Rosim Company, Brobekkveien 80, 0582, Oslo, Norway.

Nicolas Martinez:

    Master student, Faculty of Science and Technology, Norwegian University of Life Sciences, 1432, Ås, Norway.

Harsha Ratnaweera:

    Professor, Faculty of Science and Technology, Norwegian University of Life Sciences, 1432, Ås, Norway.

# 1. Introduction

In many cities, wastewater treatment plants (WWTP) are suffering from overflow induced by inefficient sewer systems and extreme weather. Straightforward sewer management measures for flood retarding purpose, such as a physical expansion of sewer pipes or construct storage units (Autixier et al. 2014; Lucas and Sample 2015) are expensive from both economic and technical-environmental perspective, especially in high wage cost countries such as Norway.

During heavy rainfalls, part of the sewer system is overloaded and another part may have unused capacity. There is a potential in utilizing the unused capacity of the sewer system itself to reduce flow to the WWTP, as well as to overflows. This approach could reduce investment, but require sophisticated modeling and management towards the different parts of the sewer system (Grum 2011; USEPA 2004; Duchesne et al. 2001).

Current studies about integrated management of WWTP and sewer system mainly focus on controlling the WWTP inflows, separate sewers and sewer expansions, together with some sustainable drainage systems (SuDS). The WWTP inflow is usually controlled by installing gates in the final collector pipe, which connects the sewer system to the WWTP. The normal function of the gates is to maximize inflow to the WWTP while keeping the water level of the final collector pipe below the plant bypass through closing or opening the gates (Maxym and Sreekanth 2015; Seggelke et al. 2005; El-Din and Smith 2002; Mark et al. 1998).

With upgraded hardware and software tools in recent years, more sophisticated strategies are becoming possible. Using a sewer hydraulic model with only one main pipe and seven secondary pipes, Carbone et al. (2014) demonstrated a decentralized control strategy that uses smart gates to exploit unused sewer capacity. The smart gates were installed downstream of the secondary pipes, before the connection to the main pipe. During heavy rainfalls, the smart gates were partially or fully closed to accumulate wastewater, thus offering an alternative solution for retarding sewer flow. However, real sewer systems are much more complex than what Carbone et al. (2014) studied, thus it is unrealistic to install control measures in all the pipes, while a simplified solution could be to apply priority control to part of the sewer system with more potential unused capacity (Perelman et al. 2012).

To manage sewer system and WWTP of the Drammen city, Norway, in a holistic way, the municipality initialized the Regnbyge 3M project. Inspired by the decentralized control solution described by Carbone et al. (2014), also considering the complexity of real sewer system, the authors proposed a "divide and conquer" strategy for Drammen city. The general technical roadmap of the "divide and conquer" strategy is: 1) divide the sewer system into several small sub-catchments, 2) grouping and recognize main features of these sub-catchments, and 3) implement management measures (conquer) according to features of sub-catchments.

The literature on water distribution system (WDS) with similar "divide and conquer" ideas may provide references to the current study. Recent literature has a significant rise in the use of machine learning. Muhammed et al. (2017) presented a new optimal rehabilitation methodology based on graph theory method, which starts with partitioning the WDS into a number of small subsystems, then pipes that might have a direct impact on system performance are identified and considered for the rehabilitation problem. Zhang et al. (2016) studied a new method for leakage identification of WDS, the method first divide a large WDS into a number of zones, then a Support Vector Machine (SVM) is trained to determine the likely leakage zones. To simplify monitoring and management of WDS, Qin et al. (2017) employed k-means clustering to group nodes with similar water quality characteristics. McKenna (2013) utilized clustering algorithm to assign multivariate water quality patterns, thus improving water quality monitoring.

Successful machine learning not only relies on the algorithms but also the datasets. Most hydrological studies only consider natural attributes of the catchments. Studies about sewer system usually only use hydraulic models (Yu et al. 2013) or monitored data (Montserrat et al. 2015). To understand different sub-catchments comprehensively, a series of natural and built factors need to be taken into consideration. Zhu et al. (2015) used factors such as the load of the sewer under different rainfall scenarios, surface characteristics of catchment and urban drainage system structure etc., to study the inundation level in different sub-catchments of the sewer system. Gaitan et al. (2016) used indices such as flood incident reports, population density, imperviousness, and catchment area to identify urban sub-catchments with higher flooding risk. Besides, interactions between sewer system and different land covers is an important issue, land coverage features have significant influences on sub-soil hydrology, thus influencing the rainfall-runoff process (Shuster et al. 2014). The quantity of the sewage varies dynamically, the response from the pervious areas depends largely on the geological conditions and that in a certain sub-catchment it has a similar proportion over time. Regulation of the inflow to the WWTP based on the sewage composition is a way to improve the performance of the WWTP (Mark et al. 1998; Gustafsson et al. 1999). The water level is also an important factor because it determines the unused space of pipes.

Dimension reduction is critical when the data used for machine learning involves time series. Wavelet transformation (WT) could decompose time series into time-frequency space; it is a powerful dimension reduction method for subsequent machine learning algorithms (Sheikholeslami 1998). WT has been used to evaluate and interpret time series data from water distribution systems (Mounce et al. 2015), improve management of CSO maintenance (Guo and Saul 2011) and water quality control (He et al. 2008). Several studies have been conducted to combine WT with machine learning algorithms. Hsu and Li (2010) proposed a WT-Self-Organizing Map (SOM) framework; this method first used WT to extract dynamic and multiscale features of the precipitation time series, and then applied SOM to identify spatially homogeneous clusters in the transformed feature space. Agarwal et al. (2016) demonstrated a WT and k-means based hybrid approach for clustering of catchments, in order to gage the complexity of a time series. The multiscale wavelet entropy is calculated based on wavelet coefficients derived from WT. This multiscale wavelet entropy extracted from wavelet analysis could reduce the dimension of the original time series data, and improve the efficiency of k-means clustering.

To the best of the authors' knowledge, only a few research works have investigated the contribution of machine learning on the management of sewer system. In the present study, machine learning is employed to undertake the task of sub-catchment grouping and feature recognition. The rest of the paper is organized as follows: section 2 describes the study area, datasets, hydraulic model and machine learning algorithms, section 3 presents grouping and feature recognition of sub-catchments, section 4 is conclusions and scopes for further researches.

## 2. Method and materials

### *2.1 Description of the study area*

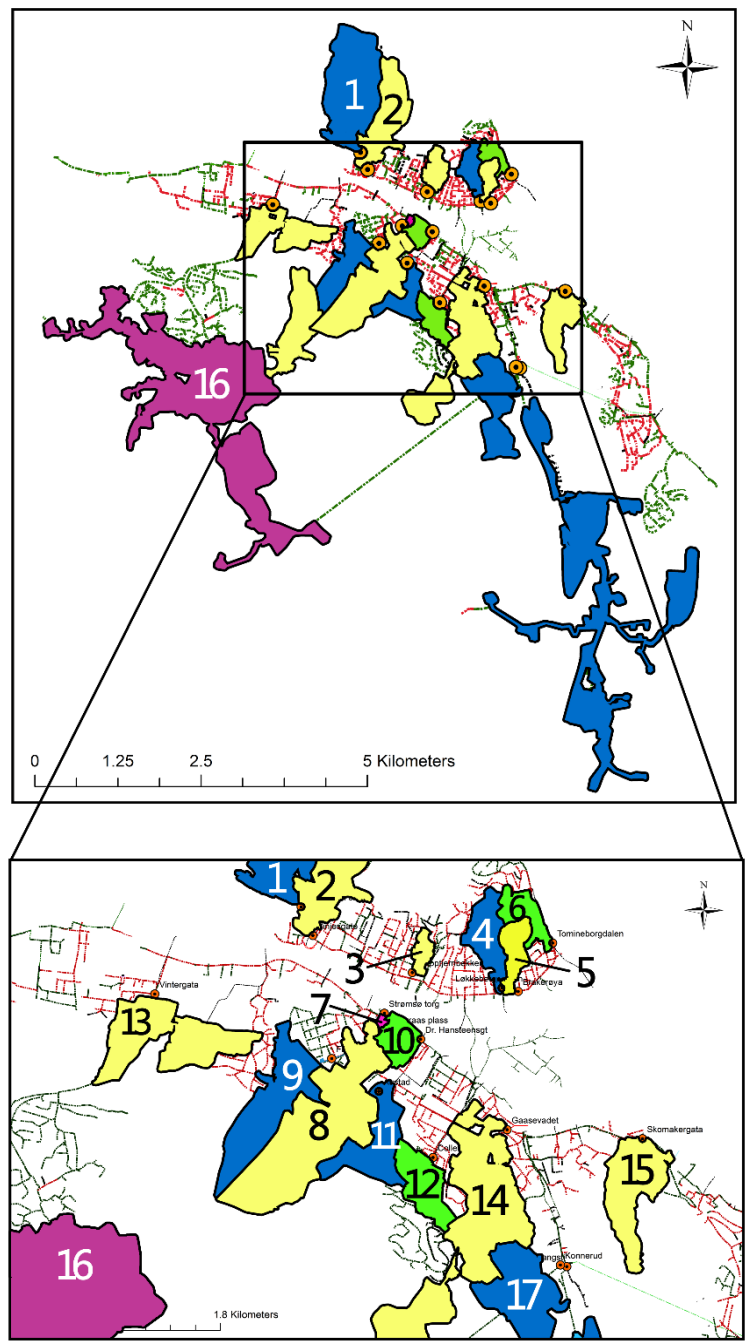

**Fig. 1.** Distribution of the sub-catchments, the numbers are sub-catchment ID

**Table 1.** Sub-catchments ID and names of sub-catchments

| Sub-catchment ID | Sub-catchment name |
|---|---|
| 1 | Børresen skole |
| 2 | Vinjesgate |
| 3 | Klopptjernbekken |
| 4 | Løkkebergparken |
| 5 | Brakerøya |
| 6 | Tomineborgdalen |
| 7 | strømsø torg |
| 8 | Torgeir vraas plass |
| 9 | Flisebekken |
| 10 | Dr.Hansteensgt |
| 11 | Austad |
| 12 | Colletts gate |
| 13 | Vintergata |
| 14 | Gaasevadet |
| 15 | Kobbervikdalen gangsti |
| 16 | Konnerud |
| 17 | Skomakergata |

The sewer system of Drammen, Norway (57.87 N, 12.66 E) has a drainage area of approximately 15 km$^2$. The total length of the sewer is around 500 km, with 43% combined sewer, 44% separate sewer and 13% storm sewer. Most parts of the sewer system is without active control, i.e. gravity based. The Solumstrand WWTP is the largest WWTP in Drammen, with the designed capacity of 130000 PE (population equivalents). Overflow is a major problem of Solumstrand WWTP.

As the first step of the "divide and conquer" strategy, the Drammen sewer system was divided into 17 sub-catchments according to its connectivity properties. The Drammen sewer system follows a tree structure. The sub-catchments distribute as branches of the tree and converge into the trunk. Sequentially, all trunks are linked to the final collector pipe, delivering wastewater to the WWTP. The sub-catchments selected in this paper are not directly connected to the other part of the sewer system, only the outlet of the sub-catchments link to the trunk. Fig. 1 and Table 1 shows distribution and names of the sub-catchments. To avoid too many Norwegian characters, all the sub-catchments will refer to their sub-catchment ID hereafter in this paper.

## 2.2 The regnbyge.no sewer surveillance system

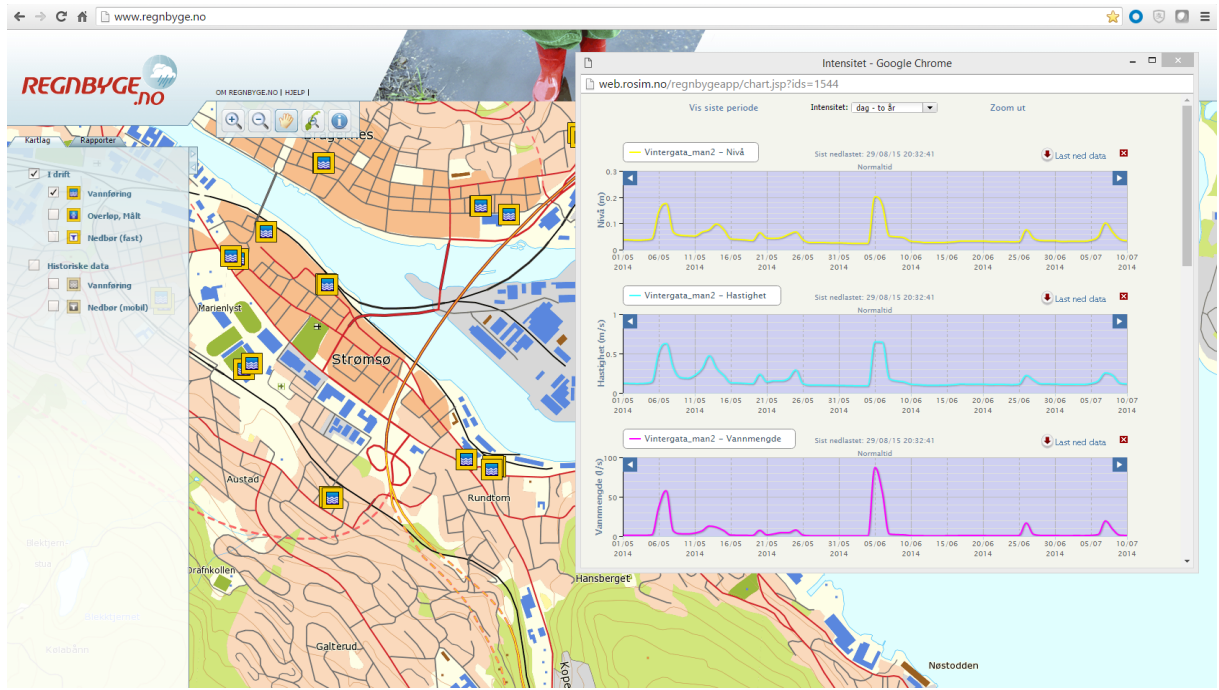

**Fig. 2.** The user interface of the regnbyge.no sewer surveillance system

In order to monitor each sub-catchment, water level sensors and velocity sensors (NIVUS GmbH; Germany) were installed in the sub-catchments' outlet pipe, connecting the sub-catchments to the trunk. The recorded water level and velocity data are transmitted wirelessly to a sewer surveillance system, regnbyge.no (http://www.regnbyge.no/) in real time. Fig. 2 displays the user interface of the regnbyge.no sewer surveillance system.

## *2.3 The Rosie model*

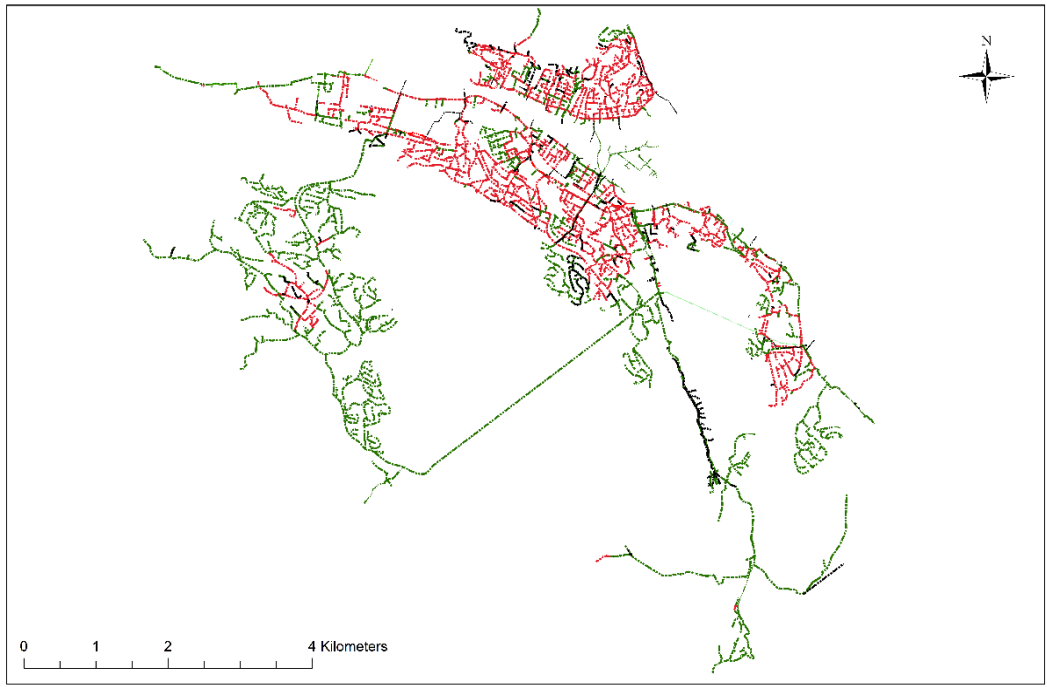

**Fig. 3.** The sewer hydraulic model for Drammen city developed by Rosie software

To simulate sewer hydrological/hydrodynamic behaviors and different control strategies, a detailed hydraulic model was developed for Drammen sewer system using Rosie. Rosie is a software developed by Rosim Company, Norway. It is an ArcGIS extension for modeling sewer systems and water distribution networks. Rosie maintains ArcGIS functionality and user interface, has several special adaptations for the Norwegian water and wastewater situations.

Rosie uses the MOUSE DHI to perform the computation. In Rosie, the direct response from the rainfall is calculated by the time–area (T-A) curve method A. The Rainfall Dependent Inflow/Infiltration (RDII) model is used to calculate the runoff generated from the pervious hydrological processes. The pipe hydrodynamic computation is based on Saint-Venant continuity and momentum equations. The MOUSE DHI RTC (real-time control) module is used to simulate sewer control. In the RTC module, the operation of control structures, such as pumps, moveable weirs or orifices, are designed by a curve that provides the relationship between water level (H) and water flow (Q). Interest readers may refer to the website of Rosie (http://web.rosim.no/index.php/tjenester/modellering-av-vann-og-avlopsnett/, *in Norwegian*) for more details. Fig. 3 shows the hydraulic model of Drammen sewer system.

## 2.4 Model calibration criteria

The hydraulic model performance was evaluated using Nash–Sutcliffe coefficient (NSE) and $R^2$, NSE is a parameter that determines the relative importance of residual variance (noise) compared to the variance in the measured data (information). The NSE is calculated by the following equation:

$$NSE = 1 - \left[\frac{\sum_{i=1}^{n}(Y_i^{obs} - Y_i^{sim})^2}{\sum_{i=1}^{n}(Y_i^{obs} - Y^{mean})^2}\right] \quad (1)$$

Where:

$Y_i^{obs}$ = the $i$-th observed data

$Y_i^{sim}$ = the $i$-th simulated data

$Y^{mean}$ = mean value of observed data

$n$ = number of observed data

NSE varies from -∞ to 1, NSE = 1 indicates a perfect correlation between simulated and observed data, values between 0.0 and 1.0 is generally acceptable. In the present research, the criteria for acceptable calibration and validation is both NSE and $R^2$ should over 0.5.

## 2.5 Description of WT

In this study, Continuous wavelet transform (CWT) was employed as a dimension reduction method. For time series f(t), CWT is the sum over all time of f(t) multiplied by the scale:

$$W_f(a,b) = |a|^{-1/2} \int_{-\infty}^{+\infty} f(t)\bar{\psi}(\frac{t-b}{a}) \, dt \quad (2)$$

Where ψ(t) is the mother wavelet, a is the parameter defining the window of analysis, b is the parameter localizing the wavelet function in the time domain, and f(t) is the complex conjugate of the basic wavelet function. $W_f(a,b)$ represents the correlation between the signal f(t) and a scaled version of the function

ψ(t). The mother wavelet is first compared with the signal at the starting point of a signal. The degree of correlation between the wavelet and the signal section is calculated. Then the mother wavelet is moved along the direction of time series data, and the process is continued until the end of time series. The mother wavelet then is updated to a larger scales and the above-mentioned process is repeated for all scales.

In this study, the mother wavelet used is the Complex Morlet wavelet. It is a widely employed mother wavelet in the field of hydrological time series studies (Li et al. 2009; Gao et al. 2014; Agarwal et al. 2016), which defined as:

$$\psi(t) = \frac{1}{\sqrt{\pi f_b}} e^{2i\pi f_c x} e^{-\frac{x^2}{f_b}} \tag{3}$$

Where $f_c$ is the wavelet center frequency, and $f_b$ is the bandwidth parameter. In this research, we defined $f_b:f_c = 1.5:1$. The main oscillation scales of time series are summarized by wavelet variance (Li et al. 2009), which defined as:

$$\mathrm{Var}(a) = \int_{-\infty}^{+\infty} |W_f(a,b)|^2 db \tag{4}$$

Where a is the frequency/scale variable, b is the time variable, $W_f(a,b)$ is the wavelet coefficient, and Var (a) is the wavelet variance. Wavelet variance interpolates the distribution of wavelet energy, the dominant oscillation periods of a time series is its extreme values.

## 2.6 The parameters for machine learning

**Table 2.** Parameters for machine learning

| Parameter groups | Parameters | Abbreviation |
|---|---|---|
| Physical attributes | Area of separate sewer (ha) | SA |
| | Percentage of green space of separate sewer (%) | SG |
| | PE associated to separate sewer | SP |
| | Area of combined sewer (ha) | CA |
| | Percentage of green space of combined sewer (%) | CG |
| | PE associated to combined sewer | CP |
| | Area of storm sewer (ha) | STA |
| | Percentage of green space of storm sewer (%) | STG |
| | PE associated to storm sewer | STP |
| | Average elevation (m) | ELE |
| Sewage composition | Total calculated sewage flow (m³) | TVO |
| | Composition of additional flow (%) | ADD |
| | Composition of ground and infiltration flow (%) | INF |
| | Composition of overland flow (%) | OVL |
| | Composition of impervious flow (%) | IMP |
| | Composition of sanitary flow (%) | SAN |
| Water level features | The first extreme value of wavelet variance | WAVE1 |
| | The second extreme value of wavelet variance | WAVE2 |
| | The third extreme value of wavelet variance | WAVE3 |
| | Mean filling degree (%) | MFD |

Parameters contain a series of natural and built factors of the sewer system was collected for the subsequent machine learning. These parameters consist of three major parts: physical attributes, sewage composition, and water level features. The physical attributes include different sewer types (combined, separate and storm), as well as the coverage of green space and PE associate to the sub-catchments. Rosie was used to calculate sewage composition because it is a kind of underlying feature that is difficult to monitor or measure (Chang et al. 2015). The Rosie classifies sewage into five components: additional flow, infiltration flow, overland flow, impervious flow and sanitary flow (foul sewage flow). One year baseline simulation was run based on rainfall data in 2014, where the portion of each component and the total calculated sewage volume was included. The parameters also include the mean filling degree of sewer water level in 2014. In order to check the long-term stability of water level, The CWT mentioned above was used to analyze the water level data under the scale of 1200 hours for water level data recorded in 2014. The top three extreme value of the wavelet variance obtained from wavelet coefficients were taken into consideration (Agarwal et al. 2016). Time series with similar significant wavelet variance share more similarities (Li et al. 2009; Gao et al. 2014).

## 2.7 Description of the machine learning algorithms

The dominant machine learning methods is used via two approaches: supervised and unsupervised. The most widely used supervised methods are SVM (Zhang et al. 2016), k-nearest neighbor, and multi-label classification (Yang et al. 2012). The most common unsupervised methods are k-means clustering (Eghbali et al. 2017), SOM clustering (Kohonen 1990), and Principal Component Analysis (PCA;

Razavi 2013). The difference of supervised methods and unsupervised methods is supervised methods require a trained set of previously labeled data.

Clustering is a very important branch of unsupervised methods as it discovers hidden structures of the data and groups similar objects into clusters (Mayer et al. 2014). Clustering is able to find the natural classes directly from the data rather than relying on labeled training data. One drawback of clustering is the data analyzer usually decides the number of clusters, because the internal structure of the data may not indicate information about an optimal number of clusters (Gauch and Whittaker 1982). The other concern is how to evaluate the quality of clustering, if similar studies existed, the quality of clustering can be evaluated by comparing its results with previous studies. Alternatively, researchers can apply several clustering algorithms to the same dataset and compare the results (Gauch 1982; Nathan and McMahon 1990). In the present study, three clustering approaches: k-means, HCA (hierarchy clustering analysis) and SOM were employed to classify the sub-catchments. Because the data have different units and scales, all the data were standardized or normalized before analysis.

### 2.7.1 K-means

The clustering algorithms can be generally divided into two groups: hierarchical clustering and partitional clustering. Usually, partitional clustering is used to fine-tune the hierarchical clustering result, via interactive relocation of points (García and González 2004). The k-means method developed by MacQueen (1967) is one of the most popular partitional clustering algorithms, it was used as a benchmark method for other clustering algorithms in this paper. The general step of the k-means clustering algorithm is:

*Initialize datasets with k points as centroids*

*Assigning each data point to its closest centroid*

*Recompute the centroid of each cluster until centroids do not change.*

In this paper, the Silhouette Coefficient (SC) (Hsu and Li 2010) was used to select the optimal number of clusters for the k-means method. The SC indicates the degree of similarity of sub-catchments within a cluster, which is defined as:

$$s(i) = \frac{b(i) - a(i)}{\max\{a(i), b(i)\}} \qquad (5)$$

Where, s(i) is the SC of sub-catchments i; a(i) is the average dissimilarity of sub-catchment i to all other sub-catchments in the same cluster, measured as Euclidean distance; and b(i) is the least average dissimilarity of sub-catchment i to the sub-catchments from any other clusters.

Then the overall quality of a clustering distribution can be measured using the formula below:

$$sc = \frac{1}{n}\sum_{i=1}^{n} s(i) \qquad (6)$$

Where n is the total number of sub-catchments. A higher value of SC indicates better discrimination among clusters of the clustering result.

### 2.7.2 HCA

HCA distribute data into a hierarchical structure according to the proximity matrix. HCA (Murtagh 1983) starts with clusters with only one object, based on distances between two objects, clusters merge from bottom to the top. The results of HCA is displayed as a tree-like diagram (dendrogram), data can be clustered by cutting the dendrogram at different branch level of the tree. As a rule of thumb, the cutting level usually between 2/3 $D_{max}$ (the maximum distance) and 1/3 of $D_{max}$ (Astel et al. 2007).

### 2.7.3 SOM

The SOM (Kohonen 1990) is a kind of artificial neural network adapted for clustering purpose. This method project the high dimensional data into a two-dimensional map. The purpose of the projection is to generate an easily readable low-dimensional representation with as much of the essential information in the input data as possible. The SOM usually have two layers, the input layer and output layer (Kohonen layer). The input layer allocates a neuron for each input variable, initialized to assign weights for each neuron, and then use the training algorithm to transfer inputs to the output layer. The output layer neurons are connected to adjacent neurons that dictates the topology of the output layer. Similar inputs should be projected close together on the output layer. Details about training the SOM is in the literature (Kohonen 1990). The SOM is particularly useful when a nonlinear mapping is inherent in the problem itself. SOM also needs to predefine the optimal number of the clusters.

## 2.7.4 PCA

PCA was used to interpolate differences and main features of the clusters obtained by clustering. Through PCA, the original high-dimensional data are projected onto lower dimensional vectors (principal components, PCs). The PCs are linear combinations of the original variables which determine the dominant patterns and the major trends in the data. The weights of each original variable when calculating the PCs called PCA loadings.

# 3. Results and discussion

## 3.1 Model calibration

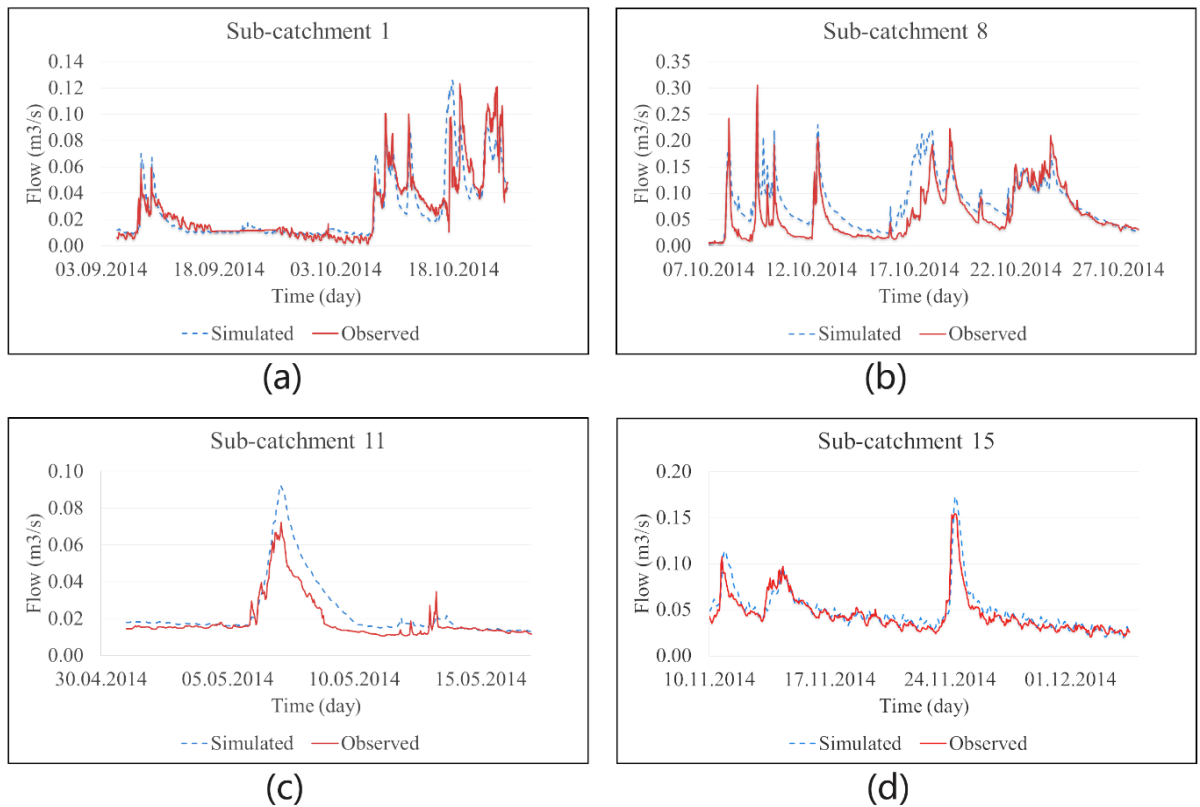

**Fig. 4.** Hydrograph of simulated and observed flows at four sub-catchments

**Table 3.** Values of NSE and $R^2$ at four sub-catchments

| Measurement point: | NSE | $R^2$ |
|---|---|---|
| Sub-catchment 1 | 0.68 | 0.71 |
| Sub-catchment 8 | 0.51 | 0.67 |
| Sub-catchment 11 | 0.54 | 0.91 |
| Sub-catchment 15 | 0.80 | 0.85 |

Fig. 4 and Table 3 shows the calibration results at four sub-catchments. The sub-catchments are from north, downtown and south part of Drammen, the period of calibration covers both dry weather season and wet weather season. The simulated curve in Fig. 4 shows a good fit for measured value, the NSE and $R^2$ in Table 3 also show acceptable values.

### 3.2 Results of WT

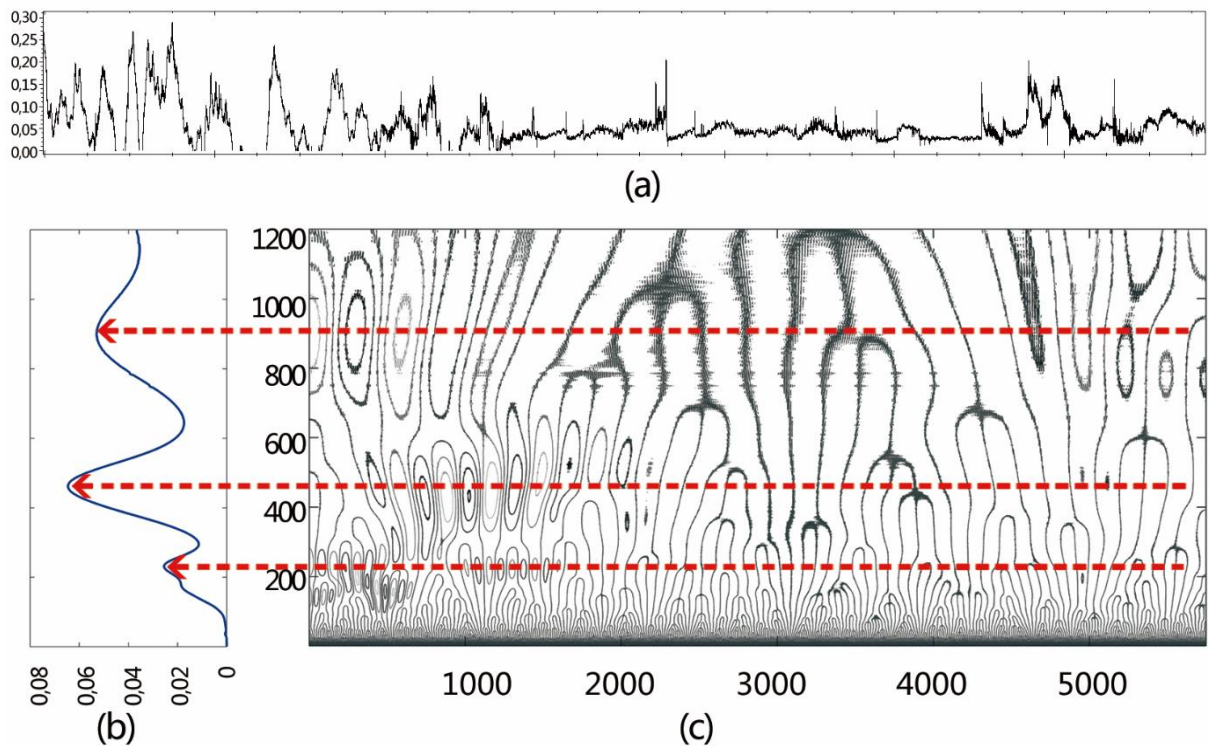

**Fig. 5.** (a) Water level time series from sub-catchment 16; (b) wavelet variance; (c) wavelet coefficient contour map

Fig. 5 shows the WT result for water level data from sub-catchment 16. Through WT, the original water level time series (Fig. 5 (a)) was transformed into a curve of wavelet variance with clear peaks (Fig. 5 (b)) that derived from the contour plot of wavelet coefficient (Fig. 5 (c)). The extreme values of wavelet variance indicate major oscillation period of the time series. The wavelet coefficient implies the intensity and the phase of the time series variation at different scales and locations, the closed contour indicates stronger oscillations. For sub-catchment 16, the wavelet variance has 150-300, 400-500, and 850-950 hours' periods, the 400-500 hours' period is significantly obvious.

### *3.3 Results of different clustering methods*

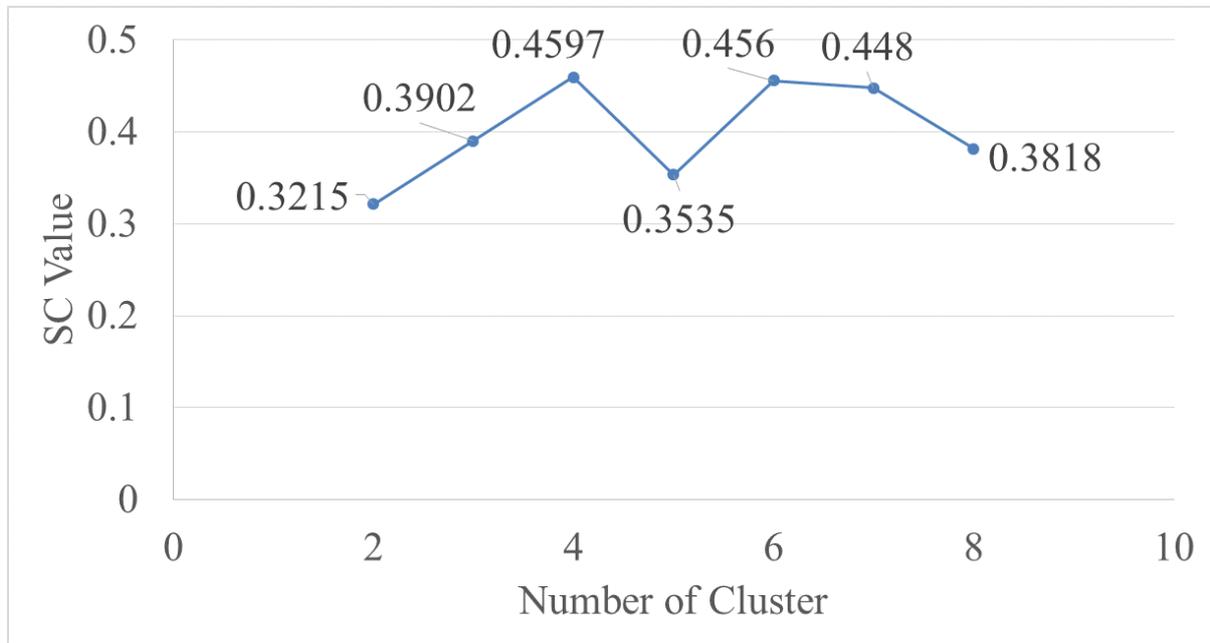

**Fig. 6.** Variation of SC values with different number of clusters for the k-means clustering algorithm

The SC values of k-means were calculated to determine the optimal number of clusters. Fig. 6 shows the variation of the SC. The most appropriate number of clusters corresponding to the highest SC value is four.

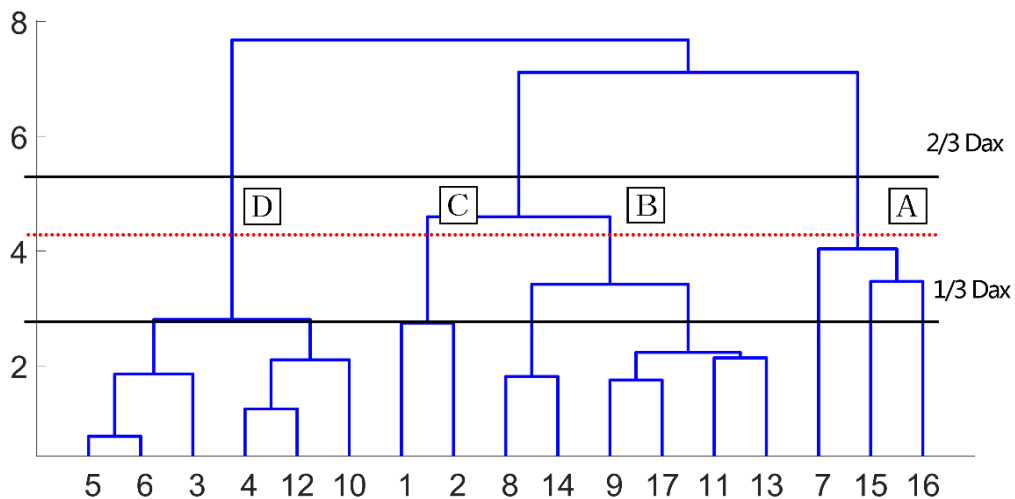

**Fig. 7.** Dendrogram of HCA with Euclidean distance and Ward's method

Fig. 7 shows the dendrogram obtained by HCA using Euclidean distance and Ward's method (Nguyen et al. 2015). Four clusters were obtained by cutting the dendrogram. The cutting level of four clusters among the suggested range (2/3$D_{max}$ and 1/3$D_{max}$).

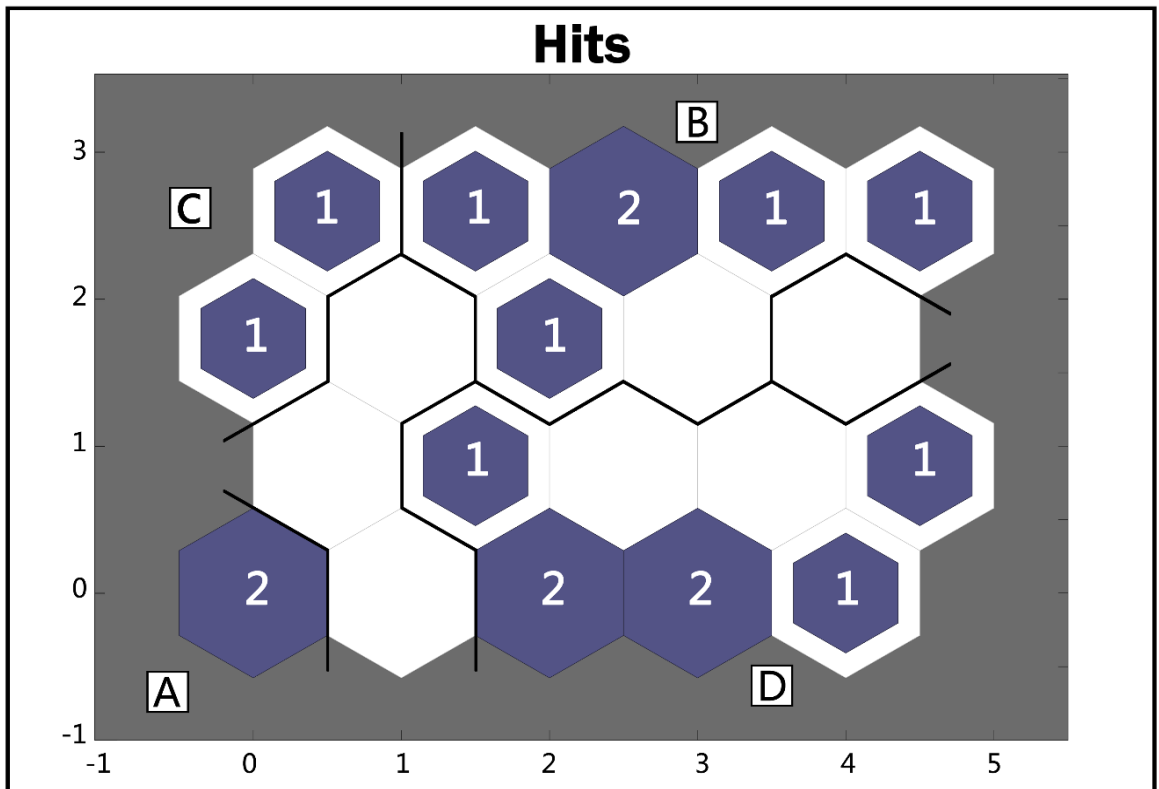

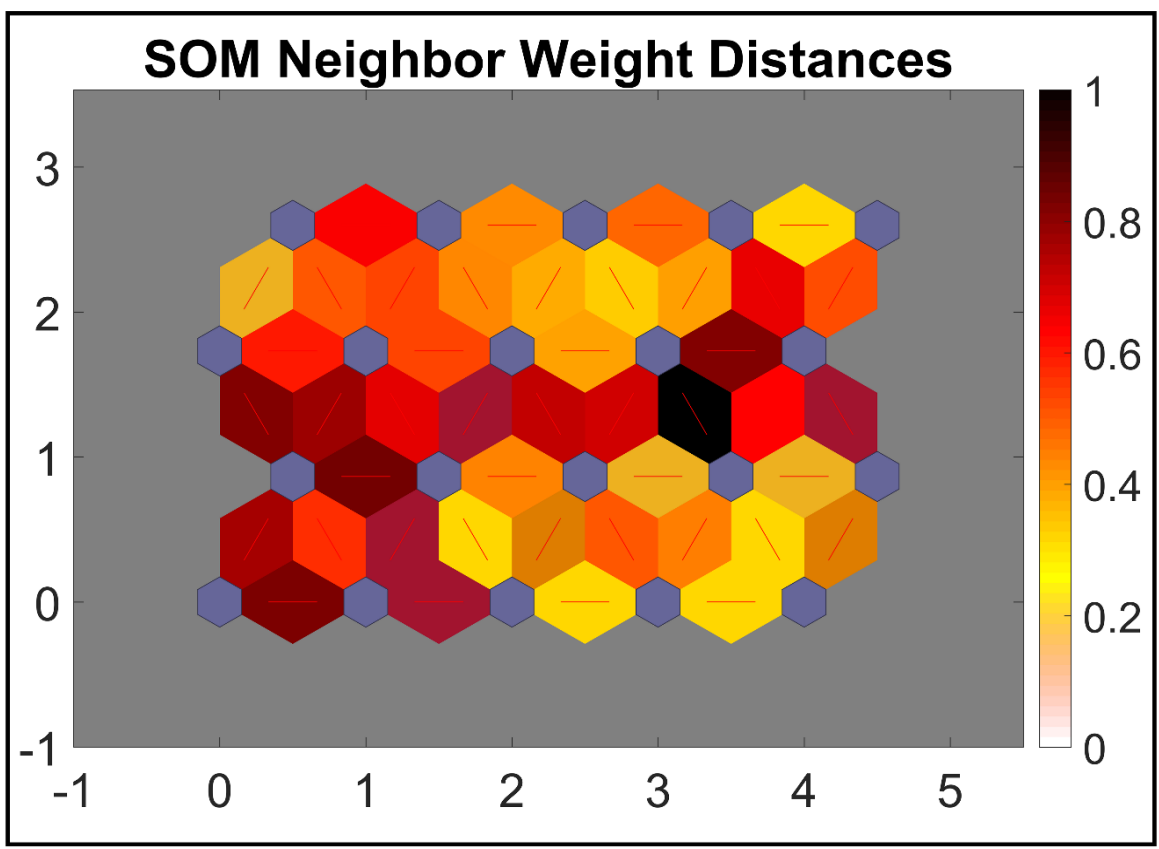

**Fig. 8.** SOM clustering result; (a) SOM hits; (b) SOM neighbor weight distances plan

The SOM in this work was selected as a hexagonal lattice with a number of nodes (n) determined using the formula: $n = 5 \times \sqrt{number\ of\ samples}$ (Astel et al. 2007). Therefore, in this paper, the SOM is $5 \times \sqrt{17} \approx 5\ columns \times 4\ rows$.

The assignment of sub-catchments to neurons was visualized by hits map (Fig. 8 (a)). The number in the middle of each hexagon indicate how many sub-catchments hit on the neurons. In the neighbor weight distances plan (Fig.9 (b)), the dark blue hexagons represent the neurons, hexagons with different colors visualize distances between neighboring neurons. The darker colors represent larger distances and vice versa. Therefore, the darker colors can be regarded as cluster boundaries, and neighboring neurons with lighter colors in between can be regarded as clusters.

As Fig. 8 (b) shows, there is a dark red and black line in the middle of neighbor weight distances plan that divides the map. In the top right part and bottom right part of the map, there are obviously two regions with yellow and orange color, so sub-catchments hit on these neurons can be grouped into two clusters. The remaining sub-catchments hits on neurons at the top left and bottom left corners. Therefore, the neighbor weight distances plan in Fig. 8 (b) revealed four clusters in the whole dataset. It is consistent with the results of k-means and HCA.

**Table 4.** Results of different clustering methods

| Catchment ID | Clustering method | | |
|---|---|---|---|
| | k-means | HCA | SOM |
| 1 | C | C | C |
| 2 | C | C | C |
| 3 | D | D | D |
| 4 | D | D | D |
| 5 | D | D | D |
| 6 | D | D | D |
| 7 | B | A | D |
| 8 | C | B | B |
| 9 | B | B | B |
| 10 | B | D | B |
| 11 | C | B | D |
| 12 | D | D | D |
| 13 | B | B | B |
| 14 | B | B | B |
| 15 | A | A | A |
| 16 | A | A | A |
| 17 | B | B | B |

*The characters A, B, C and D is assigned cluster of the sub-catchments*

The results of the three kinds of clustering methods are listed in Table 4. High consistency can be found existing among different clustering results. It reveals the reliability of clustering. In other words, it also indicates that the collected parameters could represent most of the catchments' features and balanced bias of different clustering methods. The result of SOM is adopted hereafter in this study as it reached the most consistent with other clustering methods.

## 3.4 Result of PCA

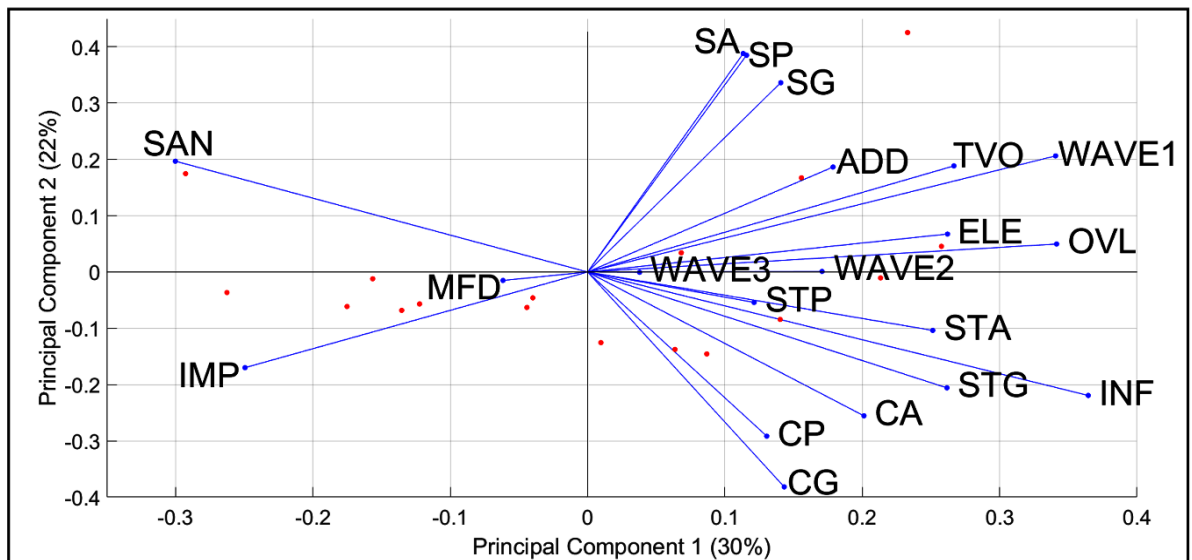

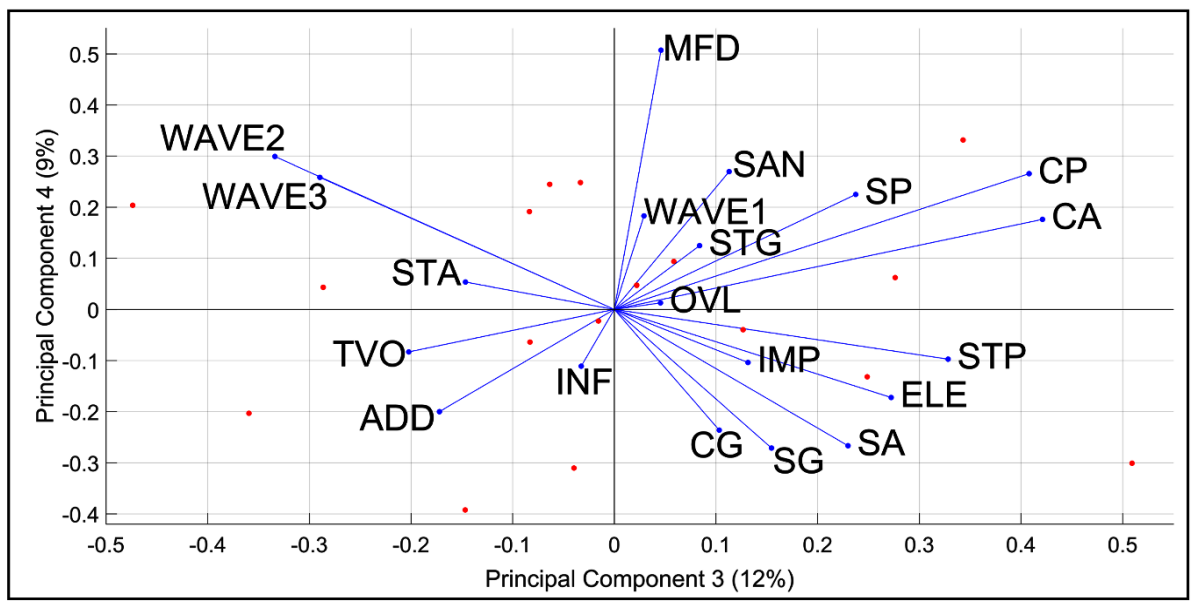

**Fig. 9.** PCA loading plots for (a) the first and second PCs and (b) the third and fourth PCs of sub-catchment attributes

**Table 5.** Values of PCA Loadings for the first four PCs of sub-catchment attributes

| Sub-catchment attributes | PC-1 | PC-2 | PC-3 | PC-4 |
|---|---|---|---|---|
| SA | 0.1133 | **0.3882** | 0.2299 | *-0.2667* |
| SG | 0.1406 | **0.3363** | 0.1546 | *-0.2712* |
| SP    | 0.1157  | **0.3851** | 0.2375   | 0.2249  |
| CA    | 0.2012  | *-0.2555* | **0.4212** | 0.1761 |
| CG    | 0.1431  | *-0.3818* | 0.1032   | *-0.2362* |
| CP    | 0.1304  | *-0.2916* | **0.4080** | 0.2656 |
| STA   | 0.2513  | -0.1037   | -0.1464   | 0.0533  |
| STG   | 0.2616  | -0.2058   | 0.0838    | 0.1246  |
| STP   | 0.1212  | -0.0540   | 0.3283    | -0.0974 |
| ELE   | 0.2620  | 0.0675    | 0.2723    | -0.1725 |
| TVO   | 0.2670  | 0.1885    | -0.2022   | -0.0832 |
| ADD   | 0.1787  | 0.1865    | -0.1720   | -0.2000 |
| INF   | **0.3645** | -0.2193 | -0.0325   | -0.1112 |
| OVL   | **0.3414** | 0.0498  | 0.0454    | 0.0128  |
| IMP   | *-0.2493* | -0.1699 | 0.1315   | -0.1040 |
| SAN   | *-0.3000* | 0.1969 | 0.1129    | 0.2696  |
| WAVE1 | **0.3407** | 0.2063 | 0.0292   | 0.1829  |
| WAVE2 | 0.1707  | 0.0009    | *-0.3339* | 0.2991  |
| WAVE3 | 0.0380  | -0.0005   | *-0.2896* | 0.2583  |
| MFD   | -0.0615 | -0.0150   | 0.0459    | **0.5072** |

*The highest positive and the lowest negative values of PCA loadings of each PCs are highlighted using bold and italics respectively*

The first four PCs explained 73% of total variances were selected to illustrate main features of different clusters and visualize correlations in the data. In Fig. 9, the direction and magnitude of the attributes indicate how each attribute contributes to the PCs. For example, it can be seen that the PC-1 is more affected by INF, OVL, and WAVE1 etc. Therefore, sub-catchments with a higher score of PC-1 is expected to be influenced more by these attributes. Attributes' loadings with similar direction and magnitude are positively correlated.

Table 5 listed values of attributes' loadings of the first four PCs. The highest positive and the lowest negative values of PCA loadings of each PCs are highlighted using bold and italics respectively. The value of loadings indicate dominate attributes of a particular PC and, therefore, can be used to interpret the PCs. It can be seen that PC1 mainly differs WAVE1 and infiltration flow to the sewer. PC2 reveals the difference between separate sewer and combined sewer. PC3 with the highest value of CA and CP, and lowest value of WAVE2 and WAVE3. PC4 has an extremely high value of MFD.

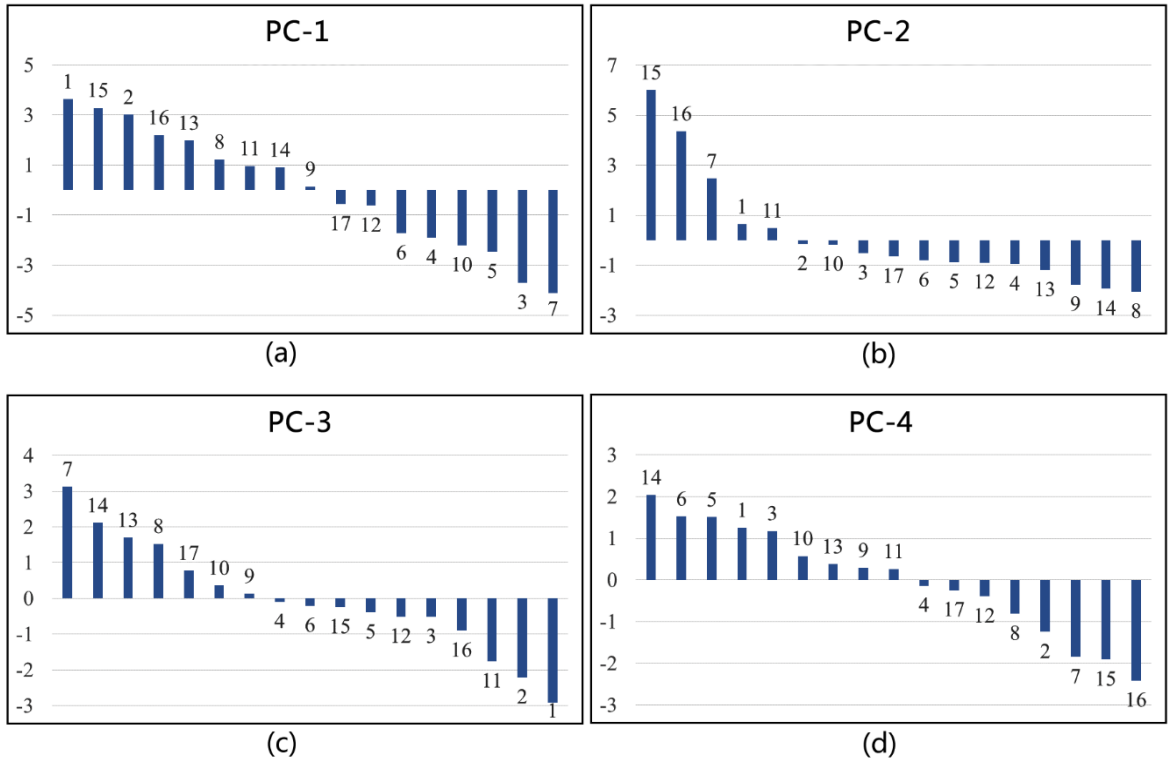

**Fig. 10.** PCA scores of sub-catchments for the first four PCs, the numbers nearby each bar are sub-catchment ID

In Fig.10, PCA scores of sub-catchments of the first four PCs are plotted. For each individual sub-catchment, the PCA score is calculated by multiply the values of PCA loadings with values of its attributes. The features of the four clusters can be interpolated as:

Cluster A: sub-catchments in this cluster have relative higher PC-1, as well as higher PC-2 and lower PC-4. Higher PC-2 means this cluster have larger area and discharges associated to the separate sewer. It also has larger green space and higher elevation. The total flow is higher because of the weights areas, and indicates of these sub-catchments are the major contributors of flow to the WWTP. The water level in these sub-catchments has a lower average mean filling degree (low PC-4). Besides, higher PC-1 reveals the larger value of WAVE1, meaning the water level has a longer time between two peak events or less fluctuate. It can be concluded that sub-catchments in cluster A are suitable for flooding retarding purpose.

Cluster B: sub-catchments in this cluster have lower PC-2 higher PC-3. The sewer system in these sub-catchments consist of both combined and separate sewer, but is more influenced by features of combined sewer (lower PC-2). Besides, the water level of these sub-catchments has more fluctuations (higher PC-3). This cluster is not ideal for control.

Cluster C: sub-catchments in this cluster have higher PC-1 and lower PC-3, which reveals the stability of the water level. A lower PC-2 compared to Cluster A implies this cluster have more combined sewer. In general, sub-catchments of this cluster are also feasible to be used as a flood retarding area but the priority is lower than Cluster A.

Cluster D: sub-catchments in this cluster have lower PC-1 and higher PC-4. Sewage in the sewer has a higher portion of impervious flow and sanitary flow (lower PC-1). As the traditional urbanized area, sewer system in this cluster are mainly combined sewer with frequent fluctuated water level (lower PC-1) and higher mean filling degree (higher PC-4). This cluster is the most unfavorable cluster for control.

### *3.5 Comparison of different control scenarios*

The clustering methods resulted in the sub-catchments being clustered into four distinct clusters. The PCA illustrated main features of the clusters, thus providing references for selection of priority control locations. The comparison of overflow volume as a function of control different clusters was performed by modeling in Rosie. A sub-catchment from each cluster was selected. The selected sub-catchments are sub-catchment 16 (from cluster A), sub-catchment 14 (from cluster B), sub-catchment 1 (from cluster C) and sub-catchment 3 (from cluster D). Regulators were introduced downstream for each outlet pipes of the selected sub-catchments, before the connection to the trunk in the hydraulic model. The regulators use a Q (flow)-H (head) relation to define the regulators control rules. The purpose of the regulators is to optimize the storage capacity of the sewer pipes upstream through accumulating or releasing the sewage.

Six scenarios were analyzed in the hydraulic model. 1 represents without control, scenario 2 represents traditional control method, i.e. only controlling the final collector pipe to the WWTP, scenario 3-6 represent applying additional control to the selected sub-catchment from Cluster A, B, C, and D respectively. To test the performance of sewer system under different control scenarios, six designed rainfall events based on Norwegian IDF (intensity-duration-frequency) curves were used. The designed rainfall events include three rainfall events with a return period of 2, 20 and 50 years, and three rainfall events represent climate change scenario with intensity 1.5 times higher than rainfall events with return periods of 2, 20 and 50 years (named 2 year plus, 20 year plus and 50 year plus). All the rainfall events have a duration of 12 hours.

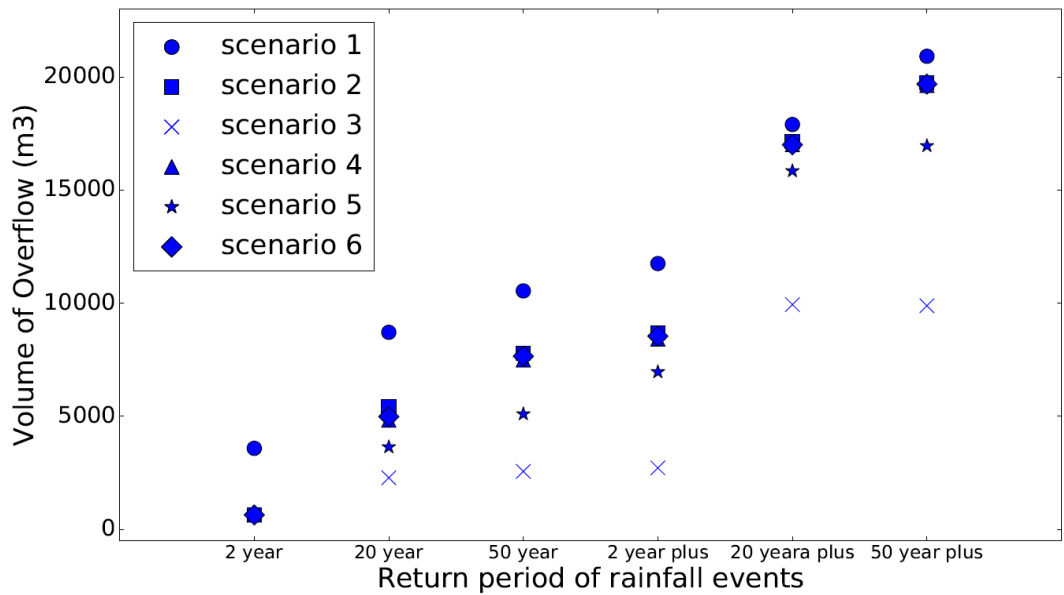

**Fig. 11.** Volume of overflow at the Solumstrand WWTP as a function of control scenarios for the six rainfall events

Fig. 11 compares the overflow volume in the six scenarios with different rainfall events. Without control, the overflow volume is particularly high. As expected, all five control scenarios led to a lower flooding volume relative to scenario 1. For traditional control scenario (scenario 2), with increased rainfall intensity, overflow reduction becomes less efficient, as shown in Fig. 11, tending to a horizontal asymptote. This suggests that only controlling the final collector pipe is not insufficient to deal with heavier rainfalls. Compared to scenario 2, the overflow reduction of scenario 4 and 6 is not distinct. There is an intermediate decrease in overflow volume for scenario 5, scenario 3 led to a substantial reduction of overflow volume, especially for extreme heavy rainfalls. The hydraulic simulations are in accordance with results of clustering and PCA, it demonstrated that priority control applied to the most suitable cluster could bring the most profitable result.

## 4. Conclusions

This paper presented a "divide and conquer" strategy to identify sub-catchments of the sewer system, which could apply priority control. The effectiveness of the proposed strategy was validated through a case study for sewer system of the Drammen city, Norway. The clustering algorithms i.e. k-means, HCA and SOM were employed to find the homogeneity of different sub-catchments. PCA was used to give insights into datasets. Results of clustering algorithms and PCA indicated the most influential attributes and highlighted the main differences between clusters, thus providing references for making a decision about prioritized control. The simulation results of hydraulic model demonstrated that priority control applied to the most suitable cluster could bring the most profitable result.

The power of WT as a dimension reduction method for subsequent clustering was demonstrated. Wavelet variance extracted from wavelet coefficient turned the original chaotic time series into clear oscillation periods. The wavelet variance could indicate long-term pipe hydraulic performance. Studies about sewer systems involve a huge amount of time series data, and the WT is a promising technique.

Drammen is a small city and only 17 sub-catchments were involved in the current study, future works can be conducted in the more complex sewer system of large cities. Another interesting direction is to enable real-time control. The hydraulic model used in the current study needs a manual operation and long computational time; this implies that smart and automatic control algorithms should be investigated.

## Acknowledgements

This work has been supported by the Regnbyge-3M project (grant number 234974), which is granted by the Oslofjord Regional Research Fund. The authors would like to thank the engineers from Rosim Company for their supports.

**References**

Agarwal, A., Maheswaran, R., Sehgal, V., Khosa, R., Sivakumar, B., and Bernhofer, C. (2016). "Hydrologic regionalization using wavelet-based multiscale entropy method." *Journal of Hydrology*, *538*, 22-32.

Astel, A., Tsakovski, S., Barbieri, P., and Simeonov, V. (2007). "Comparison of self-organizing maps classification approach with cluster and principal components analysis for large environmental data sets." *Water Research*, *41*(19), 4566-4578.

Autixier, L., Mailhot, A., and Bolduc, S., Madoux-Humery, A. S., Galarneau, M., Prévost, M., and Dorner, S. (2014). "Evaluating rain gardens as a method to reduce the impact of sewer overflows in sources of drinking water." *Science of the Total Environment*, *499*, 238-247.

Carbone, M., Garofalo, G., and Piro, P. (2014). "Decentralized Real Time Control in Combined Sewer System by Using Smart Objects." *Procedia Engineering*, *89*, 473-478.

Chang, T. J., Wang, C. H., and Chen, A. S. (2015). "A novel approach to model dynamic flow interactions between storm sewer system and overland surface for different land covers in urban areas." *Journal of Hydrology*, *524*, 662-679.

Duchesne, S., Mailhot, A., Dequidt, E., and Villeneuve, J. P. (2001). "Mathematical modeling of sewers under surcharge for real time control of combined sewer overflows." *Urban Water*, *3*(4), 241-252.

Eghbali, A., Behzadian, K., Hooshyaripor, F., Farmani, R., and Duncan, A. (2017). "Improving Prediction of Dam Failure Peak Outflow Using Neuroevolution Combined with K-Means Clustering". *Journal of Hydrologic Engineering*, 22(6), 04017007.

El-Din, A. G., and Smith, D. W. (2002). "A neural network model to predict the wastewater inflow incorporating rainfall events." *Water research*, *36*(5), 1115-1126.

Gaitan, S., van de Giesen, N. C., & ten Veldhuis, J. A. E. (2016). Can urban pluvial flooding be predicted by open spatial data and weather data?. *Environmental Modelling & Software, 85*, 156-171.

Gao, P., Geissen, V., Temme, A. J., Ritsema, C. J., Mu, X., and Wang, F. (2014). "A wavelet analysis of the relationship between Loess Plateau erosion and sunspots." *Geoderma*, *213*, 453-459.

García, H. L., and González, I. M. (2004). "Self-organizing map and clustering for wastewater treatment monitoring." *Engineering Applications of Artificial Intelligence*, *17*(3), 215-225.

Gauch Jr, H. G., and Whittaker, R. H. (1981). "Hierarchical classification of community data." *The Journal of Ecology*, 537-557.

Gauch, H. G. (1982). *Multivariate analysis in community ecology* (No. 1). Cambridge University Press.

Grum, M., Thornberg, D., Christensen, M. L., Shididi, S. A., and Thirsing, C. (2011). "Full-scale real-time control demonstration project in Copenhagen's largest urban drainage catchments." *Proceedings of the 12th international conference on urban drainage*.

Guo, N., and Saul, A. J. (2011). *Improving the operation and maintenance of CSO structures*. University of Sheffield.

Gustafsson, L. G., Hernebring, C., and Hammarlund, H. (1999). "Continuous Modelling of Inflow/Infiltration in Sewers with MouseNAM 10 years of experience." *Third DHI Software Conference*.

He, L., Huang, G. H., Zeng, G. M., and Lu, H. W. (2008). "Wavelet-based multiresolution analysis for data cleaning and its application to water quality management systems." *Expert Systems with Applications*, *35*(3), 1301-1310.

Hsu, K. C., and Li, S. T. (2010). "Clustering spatial–temporal precipitation data using wavelet transform and self-organizing map neural network." *Advances in Water Resources*, *33*(2), 190-200.

Kohonen, T. (1990). The self-organizing map. *Proceedings of the IEEE*, *78*(9), 1464-1480.

Li, C. H., Yang, Z. F., Huang, G. H., and Li, Y. P. (2009). "Identification of relationship between sunspots and natural runoff in the Yellow River based on discrete wavelet analysis." *Expert Systems with Applications*, *36*(2), 3309-3318.

Lucas, W. C., and Sample, D. J. (2015). "Reducing combined sewer overflows by using outlet controls for Green Stormwater Infrastructure: Case study in Richmond, Virginia." *Journal of Hydrology*, *520*, 473-488.

MacQueen, J. (1967). "Some methods for classification and analysis of multivariate observations." *Proceedings of the fifth Berkeley symposium on mathematical statistics and probability*., Vol. 1, No. 14, 281-297.

Mark, O., Hernebring, C., and Magnusson, P. (1998). "Optimisation and control of the inflow to a wastewater treatment plant using integrated modelling tools." *Water science and technology*, *37*(1), 347-354.

Maxym L. and Sreekanth L. (2015). Methodology to Develop Optimum Control Strategies: Controlling Wastewater Plant Inflows. *ISAWater / Wastewater and Automatic Controls Symposium, Orlando, Florida, USA*

Mayer, A., Winkler, R., and Fry, L. (2014). "Classification of watersheds into integrated social and biophysical indicators with clustering analysis." *Ecological Indicators*, *45*, 340-349.

McKenna, S., Vugrin, E., Hart, D., and Aumer, R. (2013). "Multivariate Trajectory Clustering for False Positive Reduction in Online Event Detection". *Journal of Water Resources Planning and Management*, 139(1), 3-12.

Montserrat, A., Bosch, L., Kiser, M. A., Poch, M., and Corominas, L. (2015). "Using data from monitoring combined sewer overflows to assess, improve, and maintain combined sewer systems." *Science of the Total Environment*, *505*, 1053-1061.

Mounce, S., Gaffney, J., Boult, S., and Boxall, J. (2015). "Automated Data-Driven Approaches to Evaluating and Interpreting Water Quality Time Series Data from Water Distribution Systems". *Journal of Water Resources Planning and Management*, 141(11), 04015026.

Muhammed, K., Farmani, R., Behzadian, K., Diao, K., and Butler, D. (2017). "Optimal Rehabilitation of Water Distribution Systems Using a Cluster-Based Technique". *Journal of Water Resources Planning and Management*, 143(7), 04017022.

Murtagh, F. (1983). "A survey of recent advances in hierarchical clustering algorithms." *The Computer Journal*, *26*(4), 354-359.

Nathan, R. J., and McMahon, T. A. (1990). "Identification of homogeneous regions for the purposes of regionalisation." *Journal of Hydrology*, *121*(1), 217-238.

Nguyen, T. T., Kawamura, A., Tong, T. N., Nakagawa, N., Amaguchi, H., and Gilbuena, R. (2015). "Clustering spatio–seasonal hydrogeochemical data using self-organizing maps for groundwater quality assessment in the Red River Delta, Vietnam." *Journal of Hydrology*, *522*, 661-673.

Perelman, L., and Ostfeld, A. (2012). "Water-Distribution Systems Simplifications through Clustering". *Journal of Water Resources Planning and Management*, 138(3), 218-229.

Qin, T., and Boccelli, D. (2017). "Grouping Water-Demand Nodes by Similarity among Flow Paths in Water-Distribution Systems". *Journal of Water Resources Planning and Management*, 143(8), 04017033.

Razavi, T., and Coulibaly, P. (2013). "Classification of Ontario watersheds based on physical attributes and streamflow series." *Journal of Hydrology*, *493*, 81-94.

Seggelke, K., Rosenwinkel, K. H., Vanrolleghem, P. A., and Krebs, P. (2005). "Integrated operation of sewer system and WWTP by simulation-based control of the WWTP inflow." *Water science and technology*, *52*(5), 195-203.

Sheikholeslami, G., Chatterjee, S., and Zhang, A. (1998). "Wavecluster: A multi-resolution clustering approach for very large spatial databases." *VLDB*., 428-439.

Shuster, W. D., Dadio, S., Drohan, P., Losco, R., and Shaffer, J. (2014). "Residential demolition and its impact on vacant lot hydrology: Implications for the management of stormwater and sewer system overflows." *Landscape and Urban Planning*, *125*, 48-56.

USEPA. (2004). "Report to Congress: impacts and control of CSOs and SSOs: Chapter 8 Technologies used to reduce the impacts of CSOs and SSOs." Washington, DC, USA

Yang, Q., Shao, J., Scholz, M., Boehm, C., and Plant, C. (2012). "Multi-label classification models for sustainable flood retention basins." *Environmental Modelling and Software*, *32*, 27-36.

Yu, Y., Kojima, K., An, K., and Furumai, H. (2013). "Cluster analysis for characterization of rainfalls and CSO behaviours in an urban drainage area of Tokyo." *Water Science and Technology*, *68*(3), 544-551.

Zhang, Q., Wu, Z., Zhao, M., Qi, J., Huang, Y., and Zhao, H. (2016). "Leakage Zone Identification in Large-Scale Water Distribution Systems Using Multiclass Support Vector Machines". *Journal of Water Resources Planning and Management*, 142(11), 04016042.

Zhu, Z., Chen, Z., Chen, X., & He, P. (2016). Approach for evaluating inundation risks in urban drainage systems. *Science of the Total Environment, 553,* 1-12.